%% file: main.tex
\documentclass[10pt,twocolumn,letterpaper]{article}

\usepackage[pagenumbers]{cvpr} 
\input{preamble}

\definecolor{cvprblue}{rgb}{0.21,0.49,0.74}
\usepackage[pagebackref,breaklinks,colorlinks,citecolor=cvprblue]{hyperref}

\def\paperID{8608}
\def\confName{CVPR}
\def\confYear{2024}

\title{
Activity-Biometrics: Person Identification from Daily Activities
}

\author{Shehreen Azad  \qquad  \qquad Yogesh Singh Rawat \\
{\tt\normalsize \hspace{-1.5cm} Shehreen.Azad@ucf.edu \hspace{1.2cm} yogesh@ucf.edu}
\\ 
{Center for Research in Computer Vision, University of Central Florida}
}

\begin{document}
\maketitle

\begin{abstract}
In this work, we study a novel problem which focuses on person identification while performing daily activities.
Learning biometric features from RGB videos is challenging due to spatio-temporal complexity and presence of appearance biases such as clothing color and background. 
We propose ABNet, a novel framework which leverages disentanglement of biometric and non-biometric features to perform effective person identification from daily activities. 
ABNet relies on a bias-less teacher to learn biometric features from RGB videos and explicitly disentangle non-biometric features with the help of biometric distortion. 
In addition, ABNet also exploits activity prior for biometrics which is enabled by joint biometric and activity learning.
We perform comprehensive evaluation of the proposed approach across five different datasets which are derived from existing activity recognition benchmarks. 
Furthermore, we extensively compare ABNet with existing works in person identification and demonstrate its effectiveness for activity-based biometrics across all five datasets. The code and dataset can be accessed at: \url{https://github.com/sacrcv/Activity-Biometrics/}  
\end{abstract}

\section{Introduction}
\label{sec:intro}

\begin{figure}[t!]
    \centering
    \includegraphics[width=\linewidth]{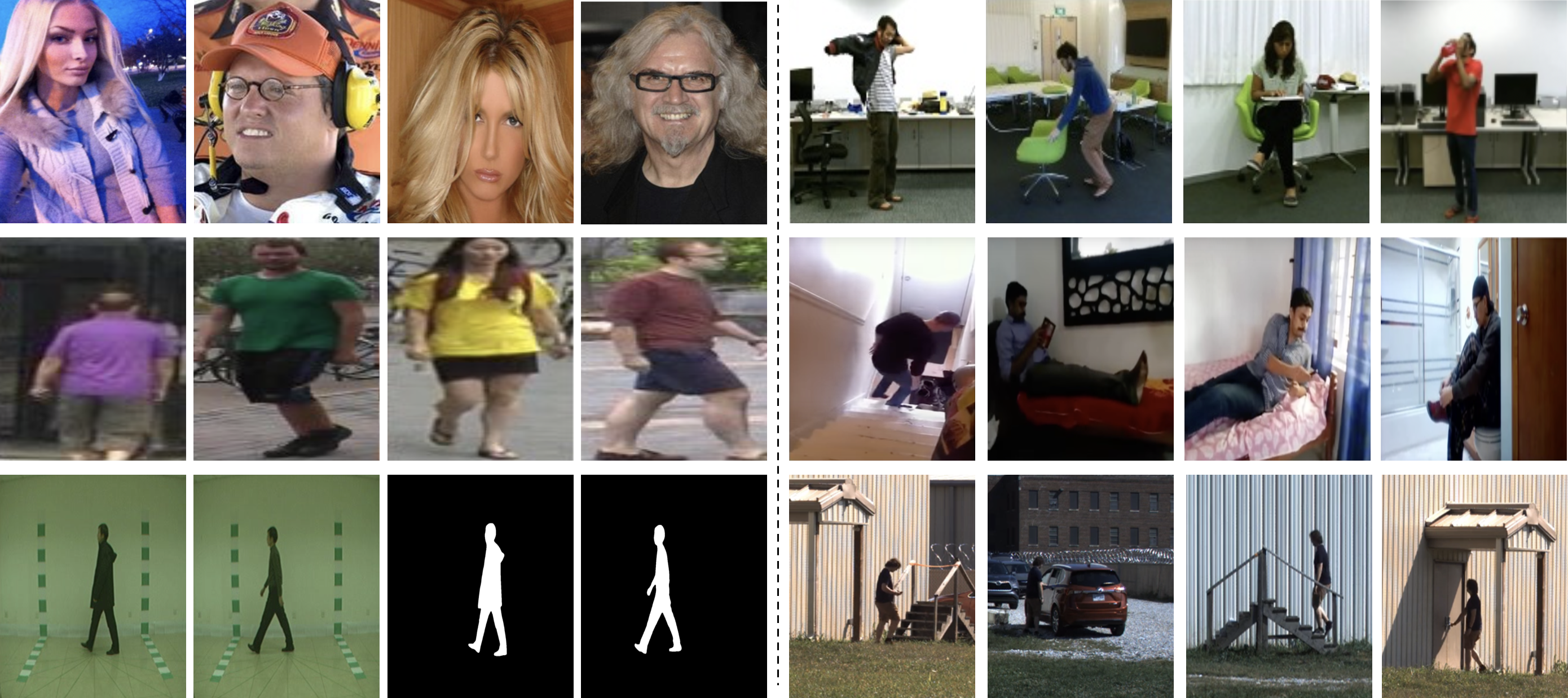}
    \caption{
    \textbf{\textit{Different approaches for person identification:}} \textit{(left)} samples for existing person identification problems such as face recognition (top: Celeb-A\cite{liu2015faceattributes}), whole body recognition (middle: Market-1501\cite{zheng2015scalable}), and gait recognition (bottom: CASIA-B\cite{yu2006framework}). (right) we focus on person identification from daily activities which presents more challenges beyond learning walking or facial patterns. We show some samples from datasets we used to study this problem; (top: NTU RGB-AB, middle: Charades-AB, bottom: ACC-MM1-Activities\protect\footnotemark).   }
    \label{fig:teaser}
\end{figure}

Person identification is an important task with a wide range of applications in security, surveillance, and various domains where recognizing individuals across different locations or time frames is essential \cite{ye2021deep}. 
We have seen a great progress in face recognition \cite{adjabi2020past, meng2021magface}, however scenarios exist where faces may not be visible, such as at long distances, with uncooperative subjects, under occlusion, or due to mask-wearing.  
This limitation prompts the exploration of whole-body-based person identification methods where most of the existing works are often restricted to image-based approaches \cite{gu2020appearance,cao2022pstr,zhang2017image}, overlooking crucial motion patterns. Video-based methods for person identification is comparatively recent area where most of the work is focused on gait recognition; mostly silhouette-based \cite{lin2021gait,fan2020gaitpart,fan2023opengait} with some recent works on RGB frames \cite{liang2022gaitedge,zhang2019gait}. However these works are mainly focused on walking style of individuals (see Figure \ref{fig:teaser}). 

\footnotetext{The subjects consented to publication}

In this work, we study a novel problem which focuses on face-restricted person identification during routine activities.
The current landscape of image-based and video-based whole-body person identification methods predominantly centers around analyzing human walking patterns from images or videos \cite{gu2022clothes,yang2023good,porrello2020robust,hou2021bicnet}. However, in real-world scenarios, the individual requiring identification might not always be engaged in walking; instead, they could be involved in various daily activities. 
It is crucial to acknowledge the significance of capturing and understanding motion cues that extend beyond simple walking patterns to ensure accurate and reliable identification in diverse and complex situations. 
These activities may offer unique cues that can prove instrumental in identifying individuals even without explicit facial information, paving the way for diverse applications in real-world scenarios, like increased surveillance in public spaces, workplace security and productivity, assistance for people requiring special needs, and smart home automation.

Learning biometrics from videos of daily activities presents several inherent challenges. 
Learning from such diverse activities amplifies the difficulty in capturing essential biometrics features. Among the crucial challenges lies the necessity to prioritize biometrics features while mitigating appearance biases present in RGB video frames, including background variations, clothing color, and other external factors. Striking a balance between extracting pertinent biometrics cues and disregarding irrelevant appearance-related biases is essential in developing robust and accurate video-based biometrics identification methods.

We propose a novel framework ABNet, which addresses some of these challenges and provides effective biometrics representation for person identification from videos of daily activities. It relies on two main components; 1) \textit{\textbf{feature disentanglement}}, and 2) \textit{\textbf{joint activity-biometrics learning}}. 
Feature disentanglement aims at avoiding appearance biases while learning the biometrics features. It explicitly learns biometrics and non-biometrics features with the help of, a) distillation from \textit{bias-less teacher}, and b) bias learning using \textit{biometrics distortion}. Joint activity-biometrics learning provides activity prior for biometrics where the knowledge of performed activity helps in person identification.

We present extensive evaluations on five different benchmarks using several metrics comparing the proposed approach with several state-of-the-art person identification methods including both image-based and video-based approaches. This comprehensive evaluation demonstrates the effectiveness and superiority of our proposed method in handling diverse datasets and scenarios for activity-based biometrics. 
Our main contributions can be summarized as, 
\begin{itemize}
    \item We study a novel problem of person identification from daily activities using RGB videos.
    \item We propose a simple and novel strategy to disentangle biometrics and non-biometrics features from videos for person identification. 
    \item We show the benefits of activity-prior for biometrics.
    \item We present several benchmarks to study this problem; these datasets are dervied from existing activity recognition datasets specifically curated for person identification. 
\end{itemize}

\section{Related work}
\label{sec:related_work}

\textbf{Image-based identification:}
Most of the existing person identification methods use image-based approach \cite{chen2021learning,hong2021fine,huang2019celebrities,yang2019person,cao2022pstr,yang2023good}. Moreover, most of these methods are designed towards learning better features in-terms of body shape, clothes, appearance etc. In recent years, learning cloth invariant features is found to be a promising direction in person identification with several works trying to address this issue. For example, one of the most popular person identification approach, CAL \cite{gu2022clothes} uses advarsarial loss to learn cloth invariant features. On the other hand, SCNet\cite{guo2023SCNet} uses a tri-stream network to learn semantically invariant features. Some works also attempt to use multiple modalities (e.g., silhouettes \cite{jin2022cloth}, skeletons \cite{qian2020long}, 3D shape \cite{chen2021learning}) etc. for better feature representation. Even though the image based methods can have better performance than some video-based methods, this performance is measured on very specific datasets, which might or might not generalize to more complex datasets where the person in consideration is performing some other activities rather than walking. 

\noindent
\textbf{Video-based identification:}
The key for video-based person identification is to extract representations robust to spatial and temporal distractors. These methods incorporate temporal information in their learned features and generally have better performance than image based methods. Several previous works \cite{dai2018video, zhang2017image} have exploited temporal cues by aggregating frames features via LSTM network. However, instead of using aggregated features extracted by RNNs, 3D CNNs perform better in terms of directly extracting spatio-temporal features that are more robust for person identification \cite{chen2020learning, liu2019spatially}. Following current research direction \cite{jiang2020rethinking, porrello2020robust,hou2021bicnet,wang2021pyramid,bai2022salient}, our work is also based on 3D CNN. 

\noindent
\textbf{Gait recognition:}
Gait recognition is a very active area of research where the goal is to identify individuals using their walking style. Existing methods mostly utilize silhouettes to avoid interference of appearance \cite{fan2023opengait,fan2020gaitpart,lin2021gait} which limits their applicability on real-world RGB videos. There are some approaches making use of RGB for gait recognition \cite{liang2022gaitedge,zhang2019gait}, but they do require silhouette in addition to RGB data.
In our proposed method we only use silhouette during training and it is not required for inference. 

\begin{figure*}[t!]
    \centering
    \includegraphics[width=\linewidth]{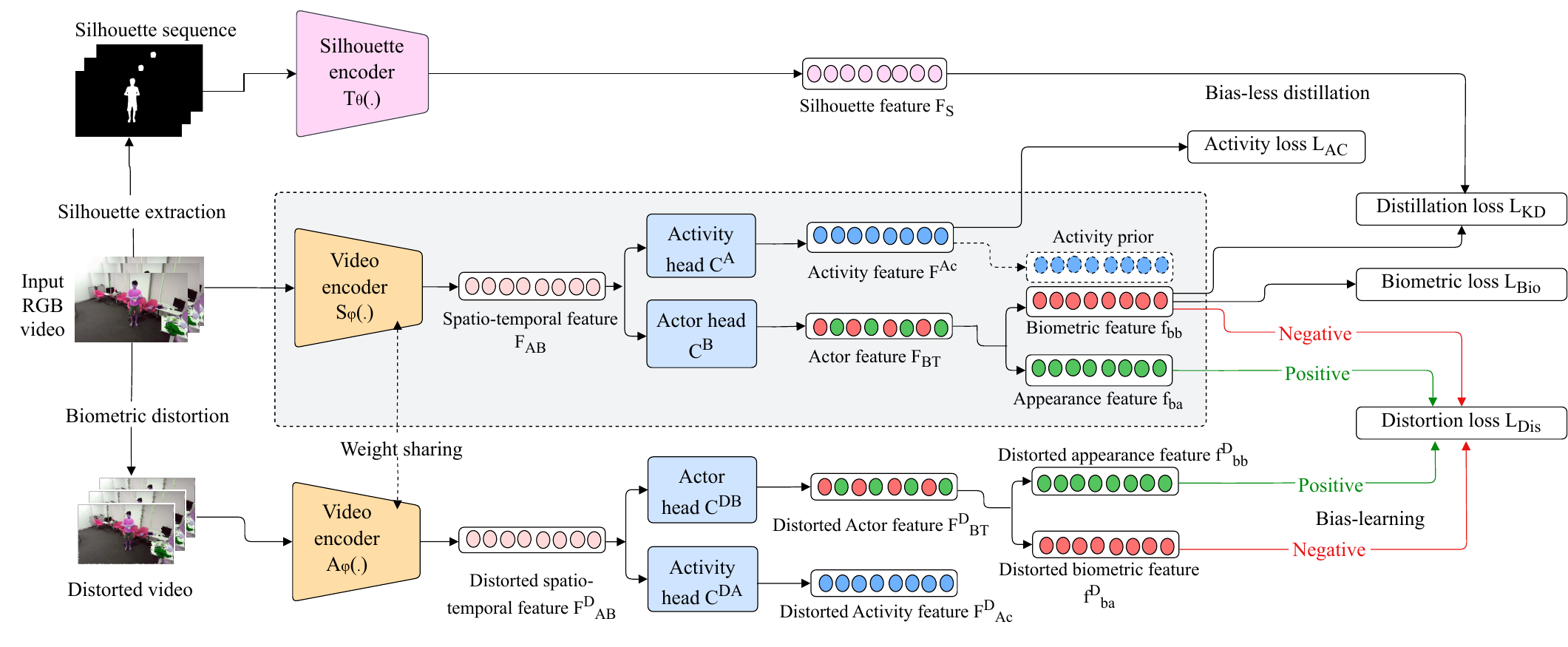}
    \caption{\textbf{\textit{Overview of our proposed method ABNet}}. RGB video is passed to a video encoder $S_\varphi (\cdot)$ for spatio-temporal feature $F_{AB}$ extraction which is passed to the activity head $C^A$ and the actor head $C^B$. $C^B$ captures both biometrics (in red) and appearance (in green) features in $F_{BT}$. To disentangle features, bias-less teacher encoder $T_\theta(\cdot)$ distills biometrics knowledge from corresponding silhouettes. The appearance feature bias is learned via a distortion network using encoder $A_\varphi (\cdot)$ on the distorted video input. Similar to $C^B$, $C^{DB}$ also captures both distorted biometrics (in red) and distorted appearance (in green) features in $F^D_{BT}$. Here, green and red denote positive and negative feature. Joint training is performed using both $C^A$ and $C^B$. During inference, only the dashed box highlighted branch is utilized.}
    \label{fig:architecture}
\end{figure*}

\noindent
\textbf{Knowledge distillation:} 
It is one of the most common techniques to transfer knowledge from a large model (teacher) to a smaller model (student) for compression and efficient learning \cite{hinton2015distilling}. 
It has also been found very effective for semi-supervised learning where the models can learn from unlabelled samples under student teacher setup \cite{tarvainen2017mean}. 
In some recent efforts it was also explored for person identification too for effective cross-view \cite{porrello2020robust} and cross-scene \cite{wu2019distilled} representation learning. 
It has been mostly explored within same modality, whereas 
we perform a cross-modal distillation  to leverage the teacher's knowledge of a different data modality to improve the performance of the student.

\section{Method}
\label{sec:method}
Our goal is to identify an individual given an RGB video of that individual performing some activity. We are using a face restricted setting to perform this task, where the face of the individual is blurred so as to avoid learning any of the facial features.
Avoiding the explicit learning of facial features is motivated by acknowledging potential issues like wearing accessory (masks, sunglasses), privacy concerns, and individuals' unwillingness to reveal their faces.

\noindent
\textbf{Problem formulation:}
Given a dataset $D$ containing elements of ${v, y^A, y^B}$ with $N$ samples, we want to train a person identification model $M$ which can provide a latent feature $F_{AB}$ for each video $v$ which can be used for matching it with the person id $y^B$. 
Here $v \in \mathbb{R}^{nXCXHXW}$ represents an RGB video, where $n$ is the number of frames, $C, H, W$ are the number of channels, height and width of the video, and $y^B$ is its ground truth actor label that is performing some activity $y^A$. 
Once trained, the model $M$ will be evaluated on a gallery $G \in {v,y^b}$ and probe $P \in {v,y^b}$. The goal is to match the id of the person $y^b$ in probe video $v$ with the correct id in videos from gallery.

\noindent
\textbf{Overview:}
We propose ABNet, Activity Biometrics Network, denoted as $M$ to solve this problem. ABNet performs \textit{biometrics-bias disentanglement} and make use of \textit{activity prior} to learn a discriminative identity feature for person identification. 
Given a video $v$, the model $M$ first extracts spatio-temporal features $F_{AB}$ with the help of a video encoder $S_{\varphi}(\cdot)$. 
The spatio-temporal feature $F_{AB}$ is split into two segments and are passed to the actor head $C^B$ for person identification as well as the activity head $C^A$ for activity recognition.
Joint biometrics and activity learning enables the use of activity-prior for biometrics.
We get actor features $F_{BT}$ from $C^B$ that contains both biometrics and appearance feature entangled with each other. Now to make the model robust to appearance bias while learning accurate biometrics features, we introduce two different components - 1) distillation from a \textit{bias-less teacher} and learning the bias using \textit{biometrics distortion}. 
The actor feature $F_{BT}$ are disentangled into biometrics feature $f_{bb}$ and appearance feature $f_{ba}$. This disentanglement for biometrics feature $f_{bb}$ is performed using distillation from a bias-less teacher $T$. On the contrary, the disentanglement for appearance feature $f_{ba}$ is done by constraining it using a distortion network $A$.
An overview of the proposed method is shown in Figure \ref{fig:architecture}.  
 
\subsection{Biometrics bias disentanglement}
Appearance bias in biometrics arises when the models overly rely on superficial visual cues, such as clothing or specific accessories for identification. This leads to challenges such as limited generalization across appearances, vulnerability to adversarial attacks, and reduced robustness to environmental variations. This bias can result in biased matching decisions, and inconsistent performance across cameras.
There has been extensive research done to avoid clothing features for person reidentification \cite{gu2022clothes,guo2023SCNet,yang2023good}, however, appearance bias can come from features other than clothes as well. To deal with this issue of appearance bias, we introduce two different aspects; 1) \textit{bias-less distillation} from a teacher network, and 2) learning the bias using negative mining through \textit{biometrics distortion}.

\noindent
\textbf{Bias-less distillation:} One split segment of the extracted feature $F_{AB}$ is fed to the actor head $C^B$, which contains $D^B_{\omega}$ that is a standard transformer decoder. We get actor feature $F_{BT}$ from $D^B_{\omega}$, which contains biometrics feature $f_{bb}$ and appearance feature $f_{ba}$. $D^B_{\omega}$ uses self-attention to process the input sequence and then projects the attention output into $f_{bb}$ and $f_{ba}$ using separate linear layers. Now to disentangle the biometrics features from the appearance features, we propose the use of silhouette features to perform bias-less distillation using teacher network $T$. $T$ is termed as bias-less because it is trained on binary silhouette video $b_s \in \mathbb{R}^{nXCXHXW}$ that corresponds to RGB video $v$, and thus have no knowledge of appearance based features. $T$ contains a silhouette encoder $T_{\theta}(\cdot)$ that takes $b_s$ as input and extracts $F_S$ features. Following \cite{hinton2015distilling} we use the standard Kullback-Leibler (KL) divergence loss to minimize the discrepancy between the probability distributions of the teacher $T$ and our model $M$.  
The distillation loss $\mathcal{L}_{KD}$ is formulated as below:
\begin{equation}
    \begin{split}
        \label{loss:kd}
        \mathcal{L}_{KD} = \tau{^2} KL(y_T || y_S),
    \end{split}
\end{equation}
where, $y_T$ and $y_S$ are the probability distribution of the teacher $T$ and our model $M$. $\tau$ is the temperature parameter that controls the softness of the teacher's output. Along with this distillation loss $\mathcal{L}_{KD}$, $C^B$ has its own biometrics loss $\mathcal{L}_{Bio}$ formulated as below: 
\begin{equation}
    \label{loss:student}
        \mathcal{L}_{Bio} = \mathcal{L}_{ce} + \mathcal{L}_{tri}, 
\end{equation}
where, $\mathcal{L}_{ce}$ and $\mathcal{L}_{tri}$ are standard triplet and cross-entropy losses for person identification formulated as below:
\begin{equation}
    \label{loss:ce}
    \mathcal{L}_{ce} = -y\log\hat{y},
\end{equation}
\begin{equation}
    \label{loss:tri}
    \mathcal{L}_{tri} =\max((D(f_a, f_p)-D(f_a,f_n)+m),0) ,
\end{equation}
where, $y$ and $\hat{y}$ are the ground truth and predicted label, $f_p$ and $f_n$ are the positive and negative features for an anchor feature $f_a$ within the same batch, $D(\cdot)$ is the Euclidean distance function, and $m$ is the margin of triplet loss. 

\begin{figure*}[t!]
    \centering
    \includegraphics[width=0.95\linewidth]{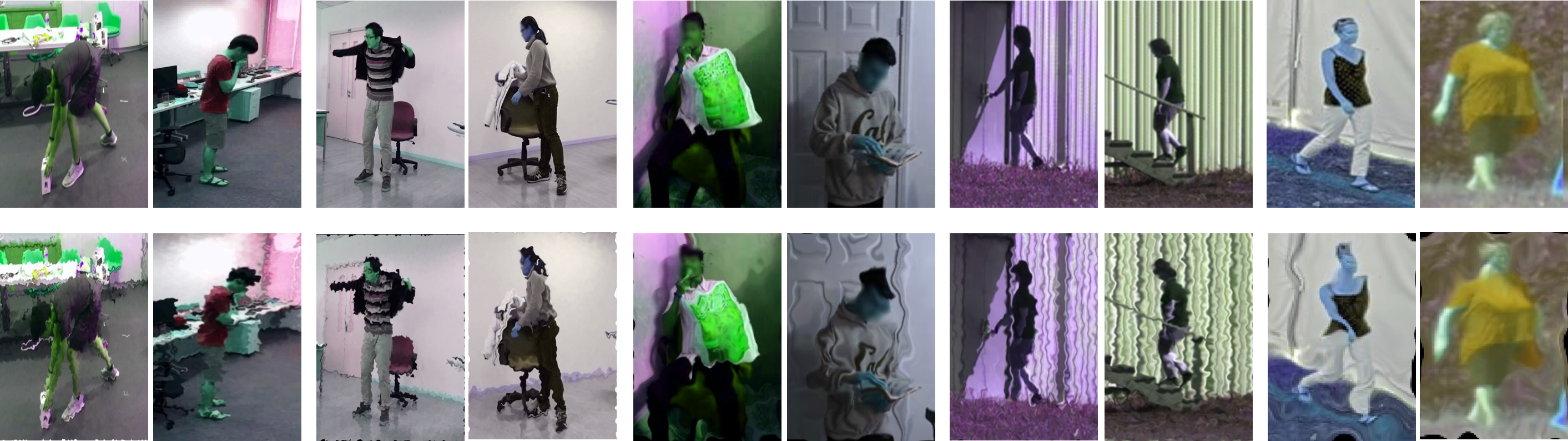}
    \caption{\textbf{\textit{Biometrics distortion:}} here original samples are shown in the top row and their corresponding distorted samples in the bottom row. From left to right, every two columns contain samples from NTU RGB-AB, PKU MMD-AB, Charades-AB, ACC-MM1-Activities and BRIAR-BGC3 dataset respectively. The subjects from BRIAR-BGC3 and ACC-MM1-Activities consented to publication.
    }
    \label{fig:distorted-sample}
\end{figure*}

\noindent
\textbf{Bias learning:}
To make the model robust to appearance bias, we introduce the distortion network $A$, which is identical to $M$ and shares weights. It contains video encoder $A_{\varphi}(\cdot)$ that takes distorted video $\hat{v} \in \mathbb{R}^{nXCXHXW}$ that corresponds to the original video $v$. 
The key idea is to distort the identity of the person while preserving the appearance. 
We rely on elastic transform \cite{elastic} 
which randomly transforms the morphology of objects in images and produces a see-through-water-like effect in the image still preserving the appearance. It is used to generate ``negative" or ``distractor" samples in the training dataset where the distorted samples will have the same appearance while changing the identity. Some sample distorted images are shown in \cref{fig:distorted-sample}. 

Similar to $M$, 
this distortion network $A$ also extracts spatio-temporal feature $F^D_{AB}$ using encoder $A_{\varphi}(\cdot)$. Since this branch is designed for bias-learning, thus the activity head $C^{DA}$ of $A$ is not utilized. On the contrary, $A$'s actor head $C^{DB}$ extracts distorted biometrics feature $f^{D}_{bb}$ and distorted appearance feature $f^{D}_{ba}$. Due to the distortion, $f_{ba}$ and $f^{D}_{ba}$ are treated as positive samples, whereas, $f_{bb}$ and $f^{D}_{bb}$ as hard negative samples. The goal is to pull together positive pairs (i.e. similar features) and push apart negative pairs (i.e. dissimilar features). We use this distorted augmentation loss $\mathcal{L}_{Dis}$ for bias learning and it is described as,
\begin{equation}
    \label{loss:distortion}
    \mathcal{L}_{Dis}= max((D(f_{ba}, f^{D}_{ba})-D(f_{bb},f^{D}_{bb})+m),0),
\end{equation}
where $D(\cdot)$ is the Euclidean distance function and $m$ is the margin for the contrastive loss. 

\subsection{Joint biometrics and activity learning}
Jointly training a network for both activity 
recognition and person identification can benefit person identification when the training data includes activities by enabling the model to learn shared representations. By learning to understand contextual cues from activities alongside actor features, the network can develop richer embeddings, thereby enhancing the model's ability to accurately identify individuals across varying activity contexts. Thus we perform joint learning of the activity and actor branch of ABNet. One segment of feature $F_{AB}$ is fed to activity head $C^A$ that contains decoder $D_\Omega^A$ that learns features $F_{Ac}$. $C^A$ is trained using $\mathcal{L}_{Ac}$ which is a standard cross-entropy loss for the activity labels regardless of the actor labels. This joint training also enables ABNet to utilize activity priors for biometrics, where we use knowledge of activity for person identification. This is accomplished by concatenating the activity features $F_{AC}$ with biometrics features $f_{bb}$ during testing.

\subsection{Overall learning objective}
Finally the model $M$ is optimized by combining all the losses which include, biometrics loss $\mathcal{L}_{Bio}$, distillation loss $\mathcal{L}_{KD}$, distortion loss $\mathcal{L}_{Dis}$ and activity loss $\mathcal{L}_{Ac}$ 
and we get the total loss $\mathcal{L}$ formulated as,
\begin{equation}
    \label{loss:overall}
    \mathcal{L} = \mathcal{L}_{Bio} + \lambda_1 \mathcal{L}_{Ac} + \lambda_2 \mathcal{L}_{KD} + \lambda_3 \mathcal{L}_{Dis} 
\end{equation}
where $\lambda_i$, $i \in [1,2,3]$ are the weights for each of the losses. 

\section{Experiments and results}
\label{sec:exp}

\textbf{Datasets:}
\label{subsec:datasets}
We perform our experiments on five different datasets which are derived from existing activity recognition benchmarks.
\textbf{1)} \textbf{NTU RGB-AB} is derived from NTU RGB+D \cite{liu2019ntu} which is a large-scale benchmark for activity recognition. 
We ignore mutual activities and consider 94 activity classes with 88692 samples fro NTU RGB-AB. 
The activity classes are divided into daily activities and medical conditions performed by a total of 106 subjects across 32 different setups, 155 different views which are shown with 3 cameras. We use the official cross-subject split for the train test separation.
\textbf{2)} \textbf{PKU MMD-AB} is derived from PKU-MMD \cite{liu2017pku} which is another large scale benchmark for activity recognition. Similar to NTU RGB-AB, we ignore mutual activities from PKU-MMD and PKU MMD-AB has 41 activity categories with almost 17,000 labeled activity instances.
These activities are performed by 66 actors in 3 different camera views and we use the official cross-subject split for our experiments. 
\textbf{3)} \textbf{Charades-AB} contains all the 9,848 annotated videos from Charades \cite{sigurdsson2016hollywood} with approximately 6.8 activities per video performed by 267 actors across 157 activity classes from a single viewpoint. We use the official train-test split for our experiments.
\textbf{4) }\textbf{ACC-MM1-Activities} \cite{brc_dataset}  is a recently curated daily activities dataset which contains 1378 annotated videos where 7 daily activities are being performed by 200 subjects from a single view-point. These activities are - enter/exit car, pull/push door, walk upstairs/downstairs, and texting. We use the official train-test split for our experiments.
\textbf{5) }\textbf{BRIAR-BGC3} \cite{cornett2023expanding} is a 
large-scale, in-the-wild person identification dataset containing samples across varying distances, environment conditions. It is mainly focused on walking/standing scenario and consists of 3 different walking conditions (structured walk, random walk and standing) performed by 1055 subjects in outdoor settings from different ranges and angle of elevation. BRIAR-BGC3 contains over 1300 hours of labeled training videos from 1055 subjects in indoor/outdoor settings. We use a 20K subset of this dataset for training with official face-restricted testing set for evaluation. \\
The videos from all five datasets undergo an arbitrarily chosen value of hue shifting. Training a model on hue-shifted data, even when appearance features are not explicitly utilized, serves to enhance the model's robustness and generalization capabilities. To facilitate face restricted person identification the faces are blurred using Gaussian blur for both the test and train split of all datasets.

\begin{table*}[t!]
    \centering
    \small
    \caption{\textbf{\textit{Comparison with state-of-the-art person identification methods}:} Evaluation shown on NTU RGB-AB, PKU MMD-AB, Charades-AB, and ACC-MM1-Activities on same-activity, View$^+$ evaluation protocol . \dag: this model was trained on silhouettes.
    }
    \begin{tabular}{c|cc|cc|cc|cc|cc}
        \hline
        \multirow{2}{*}{}&\multirow{2}{*}{Methods} & \multirow{2}{*}{Venue} & \multicolumn{2}{c|}{NTU RGB-AB} & \multicolumn{2}{c|}{ PKU MMD-AB} &   \multicolumn{2}{c|}{Charades-AB} & \multicolumn{2}{c}{ACC-MM1-Activities}  \\
        & & &Rank 1 & mAP & Rank 1 & mAP & Rank 1 & mAP & Rank 1 & mAP \\ 
        \hline
        \multirow{4}{*}{Image}& CAL \cite{gu2022clothes}&CVPR22 &73.79&28.40&81.31&49.45&43.84&25.81&69.83&42.81\\
        &PSTR \cite{cao2022pstr}&CVPR22 &69.14&34.14&84.33&47.52&37.15&24.69&57.41&34.48\\
        &SCNet \cite{guo2023SCNet}&ACM MM23 &69.89&31.47&79.53&43.55&31.73&21.89&64.68&39.79\\
        &AIM \cite{yang2023good}&CVPR23 &71.37&35.41&\underline{82.52}&48.89&40.13&28.31&74.79&49.14\\
        \hline
        \multirow{7}{*}{Video}&TSF \cite{jiang2020rethinking}&AAAI20&71.79&31.80&76.43&37.50&35.38&21.89&49.41&29.73\\
        &VKD \cite{porrello2020robust}&ECCV20 &67.41&35.63&78.35&38.54&36.31&20.71&55.38&29.57\\ 
        &BiCnet-TKS \cite{hou2021bicnet}&CVPR21 &72.71&34.45&80.79&38.52&40.31&27.34&60.44&32.79\\
        &STMN \cite{eom2021video}&ICCV21&72.98&35.08&76.55&47.92&38.72&24.49&59.44&39.68\\
        &PSTA \cite{wang2021pyramid}& ICCV21&67.41&34.78&77.44&\underline{50.42}&42.89&28.32&71.41&\underline{50.31}\\
        &SINet \cite{bai2022salient}&CVPR22 &69.41&30.68&79.58&40.80&40.31&26.90&65.39&45.41\\
        &Video-CAL \cite{gu2022clothes}&CVPR22&\underline{75.49}&\underline{39.86}&79.59&49.42&\underline{43.91}&\underline{28.51}&\underline{77.48}&50.08\\
        \hline
        \multirow{3}{*}{Baselines}& GaitGL \cite{lin2021gait} \dag &-&61.51&28.89&65.38&33.78&18.43&6.81&39.41&18.51\\
        &ResNet3D-50 \cite{he2016deep}& - &64.23 & 26.89&69.70&32.64&32.25&17.42&44.31&22.54\\
        & MViTv2 \cite{li2022mvitv2}&-&63.87&26.41&68.37&28.52&28.51&15.39&40.59&21.52\\
         \hline
         &ABNet (ours) &- &\textbf{78.76}&\textbf{40.31}&\textbf{86.83}&\textbf{57.31}&\textbf{45.84}&\textbf{31.58}&\textbf{80.43}&\textbf{52.71}\\
         \hline    
    \end{tabular}
    \label{tab:sota}
\end{table*}

\noindent
\textbf{Implementation and training details:}
The proposed method is implemented using Pytorch. 
We use ResNet3D-50 \cite{he2016deep} as the backbone of the video encoder $S_{\varphi}(\cdot)$ and 
GaitGL \cite{lin2021gait} for the teacher's silhouette encoder $T_{\theta}(\cdot)$.
The silhouettes of the RGB videos are extracted using Mask2Former \cite{cheng2021mask2former} to use as input to $T_{\theta}(\cdot)$.
 We create RGB video clips from each original video by randomly selecting 8 frames with a stride of 4. Every input frame undergoes resizing to dimensions of $256X128$.  
We train the model with a batch size of 32 with each batch containing 8 person and 4 clips for each person. Adam \cite{kingma2014adam} is used as the optimizer with weight decay of $5x10^{-4}$ and learning rate of $3.5X10^{-4}$. The model is trained for 150 epochs with a decay factor $0.1$ after every 40 epochs. The triplet loss margin $m$ is set to $0.3$ and $\lambda_i$, $i \in [1,2,3]$ in \cref{loss:overall} is set to $0.01$. During inference the activity feature $F_{Ac}$ is concatenated with the biometrics feature $f_{bb}$ that acts as the activity prior.  

\noindent
\textbf{Evaluation protocol:}
For all datasets except BRIAR-BGC3, we randomly split the test set into gallery and probe (more details in supplementary).
We use two different evaluation protocols; 1) same activity inclusive, and 2) cross-activity. For the first one, we use all the activities in the gallery whereas in cross-activity we exclude the activity in the probe while retrieval. 
Similarly, we also evaluate for same-view (View$^+$) and cross-view (View$^-$) for NTU RGB-AB and PKU MM-AB where view information is available. 
For BRIAR-BGC3, we use the official protocol for face-restricted evaluation. 

\noindent
\textbf{Evaluation metrics:}
For a thorough assessment of the model's performance, we employ rank 1 accuracy, rank 5 accuracy, mean average precision (mAP), and TAR @ 0.1\% FAR. While the first three evaluation metrics are more popular to evaluate a person identification model, the latter metric is also crucial to check the model's ability to minimize the false acceptance rate. 

\noindent
\textbf{Baseline methods:}
We consider ResNet3D-50 \cite{he2016deep}, MViTv2 \cite{li2022mvitv2} and GaitGL \cite{lin2021gait} as baselines. To further demonstrate the effectiveness of our model, we compare it against several state-of-the-art image based (CAL \cite{gu2022clothes}, PSTR \cite{cao2022pstr}, SCNet \cite{guo2023SCNet} and AIM \cite{yang2023good}) and video based (TSF \cite{jiang2020rethinking}, VKD \cite{porrello2020robust}, BiCnet-TKS  \cite{hou2021bicnet}, STMN \cite{eom2021video}, PSTA \cite{wang2021pyramid}, SINet \cite{bai2022salient},Video-CAL\cite{gu2022clothes}) person identification methods.

\subsection{Results}
In Table \ref{tab:sota}, we present rank 1 accuracy and mAP metrics for different baselines and state-of-the-art person identification methods across NTU RGB-AB, PKU MMD-AB, Charades-AB, and ACC-MM1-Activities datasets, using the same activity View$^+$ evaluation protocol. ABNet consistently outperforms both the best SOTA models and baselines across all four datasets. Table \ref{tab:sota_briar} compares ABNet with top-performing identification methods and baselines on the BRIAR-BGC3 dataset.

For a detailed evaluation, Table \ref{tab:comprehensive_eval} shows ABNet's performance across NTU RGB-AB, PKU MMD-AB, Charades-AB, and ACC-MM1-Activities datasets. This includes both same activity and cross activity evaluation protocols, featuring View$^+$ and View$^-$ settings for NTU RGB-AB and PKU MMD-AB. As view information is unavailable for Charades-AB and ACC-MM1-Activities datasets, the evaluation focuses solely on same and cross activity protocols.

\begin{table*}[t!]
    \centering
    \small
    \caption{\textbf{\textit{Comprehensive performance evaluation of ABNet:}} results shown on NTU RGB-AB, PKU MMD-AB, Charades and ACC-MM1-Activities. We observe that cross-view and cross-activity setup is the most challenging with some performance drop when compared with same activity and same view setup.}
    \begin{tabular}{c|c|cc|cc|cc|cc}
        \hline
        \multirow{2}{*}{Dataset} & \multirow{2}{*}{Evaluation Protocol} & \multicolumn{2}{c|}{R@1} & \multicolumn{2}{c|}{R@5} & \multicolumn{2}{c|}{mAP}& \multicolumn{2}{c}{TAR @ 0.1\% FAR}\\
        &&View$^+$&View$^-$&View$^+$&View$^-$&View$^+$&View$^-$&View$^+$&View$^-$\\
        \hline
        \multirow{2}{*}{NTU RGB-AB}&Same activity&78.76&77.81&85.31&82.41&40.31&38.80&39.83&35.68\\
        &Cross activity&77.01&76.43&81.37&80.37&37.64&36.14&34.92&33.79\\
        \hline
        \multirow{2}{*}{PKU MMD-AB}&Same activity&86.83&81.41&91.37&87.73&57.31&51.74&42.79&40.31\\
        &Cross activity&81.44&79.41&89.31&84.83&51.79&46.30&37.31&34.38\\
        \hline
        \multirow{2}{*}{Charades}&Same activity&45.84&-&51.04&-&31.58&-&25.39&-\\
        &Cross activity&44.82&-&52.01&-&28.78&-&22.61&-\\
        \hline
        \multirow{2}{*}{ACC-MM1-Activities}&Same activity&80.43&-&89.31&-&52.71&-&43.72&-\\
        &Cross activity&68.31&-&76.39&-&38.83&-&35.32&-\\
        \hline
    \end{tabular}
    \label{tab:comprehensive_eval}
\end{table*}

\begin{table}[t!]
    \centering
    \small
    \caption{\textbf{\textit{Performance comparison on BRIAR-BGC3}} against best state-of-the-art person identification and baselines.}
    \begin{tabular}{c|ccc}
        \hline
        Model &R@1&mAP&TAR@ 0.1\%FAR \\
        \hline
        Image-CAL \cite{gu2020appearance}&30.57&\iffalse\textbf{53.17}&\fi17.44&25.38\\
        Video-CAL \cite{gu2020appearance}&28.32&\iffalse52.44&\fi15.43&24.16\\
        PSTA \cite{wang2021pyramid}&27.75&\iffalse47.88&\fi13.78&21.54\\
        \hline   GaitGL\cite{lin2021gait}&12.61&\iffalse24.40&\fi9.51&6.44\\
        ResNet3D-50\cite{he2016deep}&22.50&\iffalse34.54&\fi12.83&19.71\\
        MViTv2\cite{li2022mvitv2}&11.78&\iffalse28.34&\fi10.21&8.44\\
        \hline
        ABNet (ours)&\textbf{34.38}&\textbf{18.78}&\textbf{26.42}\\
        \hline
    \end{tabular}
    \label{tab:sota_briar}
\end{table}

\noindent
\textbf{Comparisons:}
From Tables \ref{tab:sota} and \ref{tab:sota_briar}, it's clear that existing methods are primarily focused on identifying individuals based on walking patterns in various settings, lacking optimization for diverse activities.
Our proposed ABNet consistently outperforms existing models
across all datasets. ABNet demonstrates approximately $2\%$ to $4\%$ higher rank 1 accuracy compared to the best existing method. This consistent superiority highlights ABNet's effectiveness in person identification across diverse activity scenarios.

In Table \ref{tab:comprehensive_eval}, ABNet shows relatively stable performance across different evaluation protocols, except for ACC-MM1-Activities, which has fewer activity classes leading to larger performance gaps. The presence of overlapping activities in Charades-AB video samples reduces its performance compared to other datasets. Despite these challenges, ABNet consistently delivers strong results. Even on the predominantly walking-focused BRIAR-BGC3 dataset, ABNet outperforms the best SOTA model by $4\%$ in rank 1 accuracy. Overall, ABNet demonstrates robust performance, particularly on datasets with diverse activity classes.

\subsection{Ablations}
To verify the effectiveness of ABNet and each of its components, we perform ablation study on the NTU RGB-AB dataset in Table \ref{tab:ablation} on the same activity evaluation protocol. Refer to the supplementary for ablation study on the cross activity evaluation protocol. Here, B/L stands for the baseline which is just the backbone model taking RGB video as input. K/D stands for bias-less distillation, A/P stands for activity prior, and lastly F/D stands for the bias learning. 

\noindent
\textbf{Effect of bias-less distillation:}
Introducing bias-less distillation, either independently (row 2) or with an activity prior (row 4), leads to notable performance improvements over the baseline. However, combining bias-less distillation and activity prior demonstrates superior performance  over independent use of distillation, showcasing their synergistic effect on model enhancement.

\noindent
\textbf{Effect of bias learning:}
Incorporating bias learning through a distorted video encoder branch boosts model performance even more (row 5). Similar to bias-less distillation, combining bias learning with an activity prior yields the best overall performance (row 6), highlighting the importance of their synergy in enhancing model robustness and disentangling biometrics and appearance information.

\noindent
\textbf{Effect of activity prior:}
Incorporating activity and biometrics features during inference significantly enhances performance compared to using only the baseline model (row 3). This integration consistently improves model efficacy across various model configurations demonstrating the role of activity recognition for biometrics.

\begin{table}[t!]
    \centering
    \caption{\textbf{\textit{Ablation studies}} of each component of ABNet on NTU RGB-AB on same activity evaluation protocol.}
    \small
    \begin{tabular}{p{0.4cm}p{0.4cm}p{0.4cm}p{0.4cm}|cc|cc}
         \hline
         \multirow{2}{*}{B/L} & \multirow{2}{*}{K/D} & \multirow{2}{*}{A/P} & \multirow{2}{*}{F/D} & \multicolumn{2}{c|}{View$^+$} & \multicolumn{2}{c}{View$^-$}\\
         &&&&R@1 & mAP& R@1 & mAP\\

         \hline
         \checkmark &&&& 64.23 & 26.89 & 62.10 & 22.45 \\
         \checkmark&\checkmark & & & 69.31 & 28.01 & 66.57 & 24.29 \\
         \checkmark && \checkmark &  & 69.43& 27.97& 67.37 &24.77\\
          \checkmark&\checkmark & \checkmark & & 72.89	&32.38& 70.17 &30.68\\
         \checkmark&\checkmark & & \checkmark & 76.70	&36.21& 73.82 &33.18\\
          \hline
         \checkmark&\checkmark & \checkmark & \checkmark & \textbf{78.76}	&\textbf{40.31}& \textbf{77.81} &\textbf{38.80}\\
         \hline
    \end{tabular}
    \label{tab:ablation}
\end{table}

\begin{figure*}[t!]
    \centering
    \includegraphics[width=0.85\linewidth]{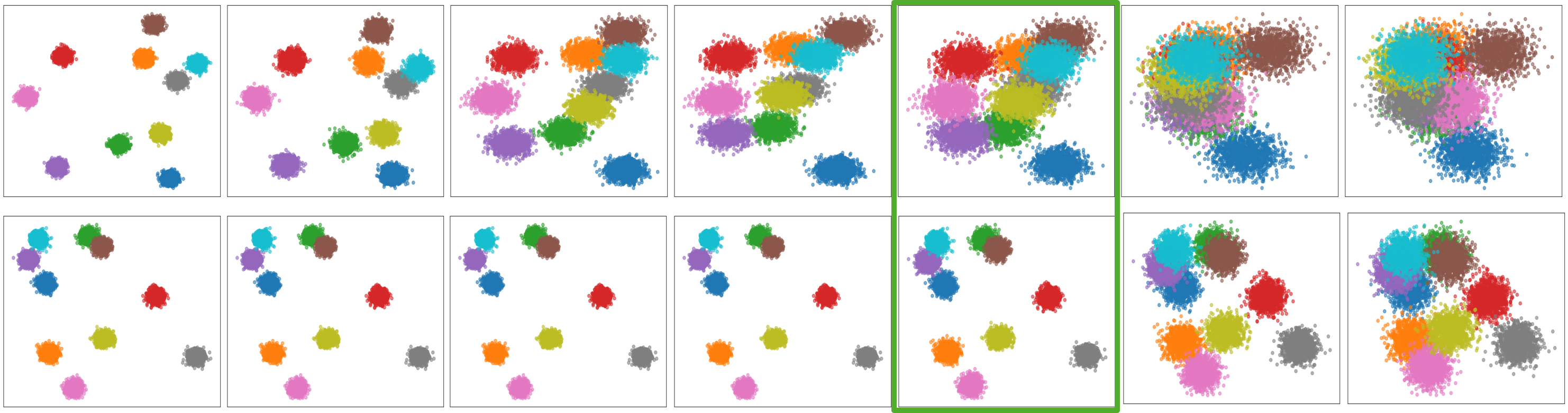}
    \caption{\textbf{\textit{Effect of distortion on feature space:}} The t-SNE plots illustrate the impact of varying distortion amount $\alpha \in [0,50,100,150,250,300,350]$ on biometrics (top) and appearance (bottom) features of ABNet for ten random NTU RGB-AB identities. As $\alpha$ increases from left to right, the optimal results occur at $\alpha=250$ (shown in square) where biometrics changes while appearance remains consistent. Beyond $\alpha=250$, appearance gets distorted too, making it unsuitable for disentanglement.
    }
    \label{fig:alpha-tsne}
\end{figure*}

\subsection{Discussion and analysis}

\textbf{Effect of distortion:}
Figure \ref{fig:alpha-tsne} presents t-SNE plots of the biometrics and appearance feature space for ten NTU RGB-AB individuals at $\alpha \in [0,50,100,150,200,250,300,350]$, where $\alpha$ represents the amount of distortion. For optimal $\alpha$ we want to find such a value where biometrics feature clusters are overlapped due to being negative, but appearance feature clusters still remain relatively same due to being positive. Increasing $\alpha$ causes more overlap in biometrics feature clusters; whereas up to $\alpha=250$, appearance feature clusters remain relatively stable. However, beyond this point, excessive distortion causes overlapping appearance clusters. Thus $\alpha=250$ is selected as the optimal value. From the quantitative results presented in Table \ref{tab:distortion} similar effect of $\alpha$ is observed on the model's performance. 

\begin{table}[t!]
    \centering
    \small
    \caption{\textbf{\textit{Effect of distortion}} on model performance for NTU RGB-AB on the same activity evaluation protocol}
    \begin{tabular}{c|cc|cc}
         \hline
         \multirow{2}{*}{Distortion amount}&\multicolumn{2}{c|}{View$^+$} & \multicolumn{2}{c}{View$^-$} \\
         & R@1 & mAP& R@1 & mAP \\
         \hline
        $\alpha=200$&	78.23&38.31&	76.81&37.91\\
        $\alpha=250$& \textbf{78.76}	&\textbf{40.31}& \textbf{77.81} &\textbf{38.80}\\
        $\alpha=300$& 75.24&31.42&	73.17&29.84\\
        \hline 
    \end{tabular}
    \label{tab:distortion}
\end{table}

\begin{figure}[t!]
    \centering
    \includegraphics[width=\linewidth]{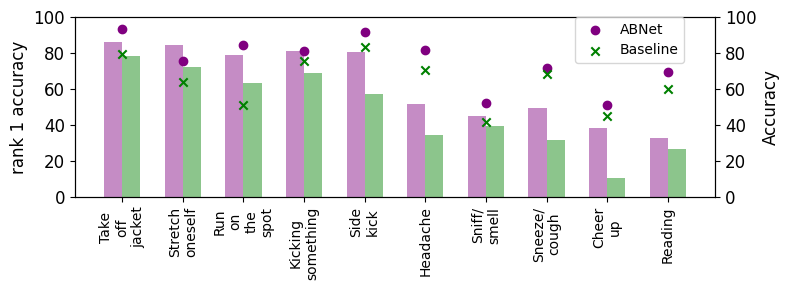}
    \caption{\textbf{\textit{Performance analysis across activities:}} The bar plot on left axis shows rank 1 identification accuracy for given activity of ABNet against baseline on 10 activity (5 best and 5 worst) classes of NTU RGB-AB. The scatter plot with markers on right axis shows activity recognition accuracy for corresponding classes.}
    \label{fig:act_per}
\end{figure}

\noindent
\textbf{Performance analysis across activities:} Figure \ref{fig:act_per} illustrates the comparison between our method and the baseline across selected activities, encompassing the top five best and bottom five worst instances in person identification performance. Notably, activities posing challenges for person identification, resulting in lower performance, also exhibit reduced accuracy in activity recognition, except for a few exceptional activity classes. This correlation underscores the consistent relationship between the difficulty of identifying individuals within activities and the corresponding accuracy of recognizing those activities.

\noindent
\textbf{Effect of face restriction:} Table \ref{tab:face_blur} illustrates the model's performance on the same activity evaluation protocol, indicating a minimal increase in performance despite the presence of facial features. This suggests the model's resilience to facial variations, showcasing its capability to identify individuals based on non-facial cues. ABNet demonstrates stability in performance even after the removal of facial appearance cues, highlighting its reliance on other distinguishing features, such as activity-related cues. 

\begin{table}[t!]
    \centering
    \small
    \caption{\textbf{\textit{Effect of face restriction}} on model performance for NTU RGB-AB on same activity evaluation protocol}
    \begin{tabular}{c|cc|cc}
         \hline
         \multirow{2}{*}{Face Restricted}& \multicolumn{2}{c|}{View$^+$} & \multicolumn{2}{c}{View$^-$} \\
         & R@1 & mAP& R@1 & mAP \\
         \hline
         Yes & 78.76	&40.31& 77.81 &38.80\\
        No& 79.24&41.64& 78.87&40.04\\
        \hline 
    \end{tabular} 
    \label{tab:face_blur}
\end{table}

\begin{figure}[t!]
    \centering
    \includegraphics[width=\linewidth]{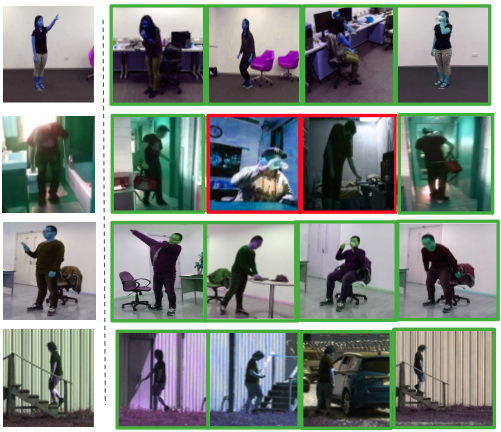}
    \caption{\textbf{\textit{Top 4 rank retrieval}} samples for ABNet on NTU RGB-AB, Charades-AB, PKU MMD-AB and ACC-MM1-Activities on row $1,2,3,4$ respectively. The left most column shows the probe and rest of the columns are the retrieved list. Accurate retrieval is shown with green box and inaccurate with red. The subjects from ACC-MM1-Activities consented to publication.}
    \label{fig:qual}
\end{figure}

\noindent
\textbf{Qualitative results:}
In addition to the quantitative results, we show top 4 rank retrieval results in Figure \ref{fig:qual}. Each row in this figure corresponds to a probe (left) and the identities retrieved (right) by ABNet. The retrieval list shows accurate person identification across a variety of activities and appearance, effectively highlighting ABNet's ability to learn from activity cues rather than appearance.

\section{Conclusion}
\label{sec:conclusion}
In this work we study a novel problem of person identification from videos of daily activities.  
We propose ABNet, a simple approach to solve this problem which relies on feature disentanglement and activity prior for person identification. 
This approach incorporates feature disentanglement at both biometric and appearance levels, leveraging distinct strategies to enhance accuracy and mitigate biases. By distilling biometric knowledge from a bias-free silhouette-trained model and learning appearance biases via elastic distortion-based transformations, our framework ensures a comprehensive understanding of individuals' inherent biometric traits while accounting for appearance variations. Moreover, the integration of an activity prior during inference further enriches the model's capabilities. Through extensive evaluations on five benchmark datasets derived from large-scale activity recognition datasets, our approach consistently surpasses several state-of-the-art methods.

\section{Acknowledgement}
\label{sec:acknowledgement}
This research is based upon work supported in part by the Office of the Director of National Intelligence (IARPA) via 2022-21102100001. The views and conclusions contained herein are those of the authors and should not be interpreted as necessarily representing the official policies, either expressed or implied, of ODNI, IARPA, or the US Government. The US Government is authorized to reproduce and distribute reprints for governmental purposes notwithstanding any copyright annotation therein.

{
    \small
    \bibliographystyle{ieeenat_fullname}
    \bibliography{main}
}
\appendix
\clearpage
\setcounter{page}{1}
\setcounter{section}{0}
\setcounter{table}{0}
\setcounter{figure}{0}
\maketitlesupplementary

\noindent
The supplementary is organized as follows:
\begin{itemize}
    \item Section \ref{sec:gall_probe} provides detailed description of the gallery probe setup of the datasets.
    \item Section \ref{sec:comparison} provides comparison of ABNet with state-of-the-art methods on the cross-activity evaluation protocol.
    \item Section \ref{sec:discuss_supple} provides ablations and some more discussion and analysis 
    \item Section \ref{sec:qualitative_supp} provides some more qualitative samples of retrieval results of ABNet.
\end{itemize}
The code and all the datasets used for this work will be made publicly available.

\section{Gallery probe setup}
\label{sec:gall_probe}
We evaluate the performance in terms of same activity and cross activity. In the same activity evaluation protocol, probe and gallery contains all the activities, however, probe contains a smaller subset of samples and the rest are placed in gallery. In the cross activity evaluation protocol, probe and gallery contains mutually exclusive activities, where probe contains a smaller subset of samples and rest of the samples from those activities are discarded; on the contrary the gallery contains all samples from a certain activity. Here for each actor there are multiple activity samples, and each activity again has different view-point or setup variation (for NTU RGB-AB and PKU MMD-AB). The samples are randomly selected for gallery and probe sets. For NTU RGB-AB and PKU MMD-AB two variations are checked - probe view included in gallery (View$^+$) and probe view excluded from gallery (View$^-$) in case of both same activity and cross activity protocol. However, since Charades and ACC-MM1-Activities does not contain multiple view points, the evaluation protocol with inclusion/exclusion of probe view from gallery is not relevant in these case. Table \ref{tab:data_description} illustrates a detailed description of all the datasets. 
\begin{table}[t!]
    \centering
    \small
    \caption{\textbf{\textit{Dataset statistics}}}
    \begin{tabular}{p{1.7cm}c|ccc}
         \hline
         Dataset& Split&\#actors&	\#activities &	\#samples\\
         \hline
         \multirow{3}{1.7cm}{NTU RGB-AB}& train	&85&\multirow{3}{*}{94}&70952\\
         & gallery& \multirow{2}{*}{21}	&	&14192\\
         &probe&&&			3548\\
         \hline
         \multirow{3}{1.7cm}{PKU MMD-AB}& train	&53&\multirow{3}{*}{41}&13634\\
         & gallery& \multirow{2}{*}{13}	&	&2727\\
         &probe&&&			681\\
         \hline
         \multirow{3}{1.7cm}{Charades-AB}& train	&214&\multirow{3}{*}{157}&45111\\
         & gallery& \multirow{2}{*}{53}	&	&9022\\
         &probe&&&			2256\\
         \hline
         \multirow{3}{1.7cm}{ACC-MM1-Activities}& train	&182&\multirow{3}{*}{7}&7717\\
         & gallery& \multirow{2}{*}{45}	&	&1543\\
         &probe&&&			386\\
         \hline
         \multirow{3}{1.7cm}{BRIAR-BGC3}& train	&870&\multirow{3}{*}{3}&20000\\
         & gallery& \multirow{2}{*}{130}	&	&4171\\
         &probe&&&			922\\
         \hline
    \end{tabular}
    \label{tab:data_description}
\end{table}

\section{Comparison with state-of-the-art methods}
\label{sec:comparison}
We present the comparison of different state-of-the-art methods against our proposed ABNet to show its effectiveness across NTU RGB-AB, PKU MMD-AB, Charades-AB and ACM-MM1-Activities datasets on the cross-activity View$^+$ evaluation protocol in Table \ref{tab:sota_ca} which corresponds to Table \ref{tab:sota} of the main paper. Similar to the quantitative comparisons presented in the main paper, in case of cross-activity evaluation protocol as well, ABNet outperforms all the existing methods and baselines by a competitive margin in terms of both evaluation metrics. This shows the robustness of our method against same or cross activity evaluation. 

\section{More analysis and discussion}
\label{sec:discuss_supple}
\begin{figure}[H]
    \vspace{-.4cm}
    \centering
    \includegraphics[width=\linewidth, height=2.4cm]{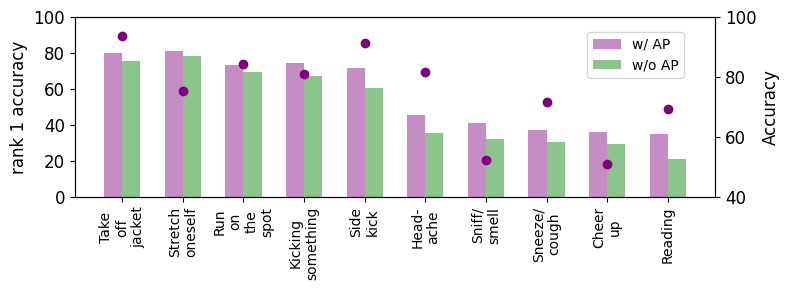}
    \caption{\textbf{\textit{Performance analysis w/ and w/o activity prior}}; bars represent biometrics rank 1 and dots represent activity accuracy.}
    \label{fig:ap}
\end{figure}

\textbf{Ablations on cross-activity evaluation protocol.} Table \ref{tab:ablation_ca} illustrates the effect of each component of our proposed ABNet on NTU RGB-AB dataset on the cross-activity evaluation protocol. This table is an extension of Table \ref{tab:ablation} of the main paper and similar to the same-activity evaluation protocol, the performance of the model remains stable in case of cross-activity and also each modification component gives a performance boost to the model, which finally contributes to the overall model's performance. Now, some activities might be easier to recognize and hence, we perform an experiment on top 5 best and top 5 worst performing activities with and without the activity prior (AP) to see whether the easily recognizable activities introduce any bias through the activity information. 
In Figure \ref{fig:ap} we see that the performance pattern remains consistent across activities with or without AP which indicates that AP consistently helps and the difficulty level of activities do not introduce any bias. 
\vspace{-.4cm} 

\begin{figure*}[t!]
    \centering
    \includegraphics[width=\linewidth,height=4cm]{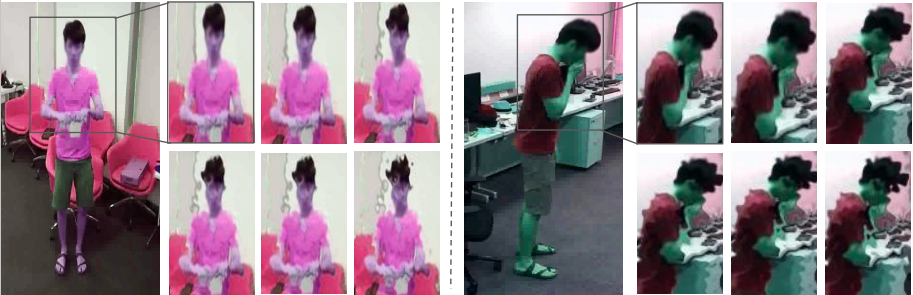}
    \caption{\textbf{\textit{Effect of distortion amount}} Original sample zoomed in to show effect of $\alpha=50,100,150$ (top) and $\alpha=200,250,300$ (bottom). As $\alpha$ increases, the distortion keeps increasing.}
    \label{fig:distortion_zoom}
\end{figure*}

\begin{table*}[t!]
    \centering
    \small
    \caption{\textbf{\textit{Comparison with state-of-the-art person identification methods}:} Evaluation shown on NTU RGB-AB, PKU MMD-AB, Charades-AB, and ACC-MM1-Activities on cross-activity, View$^+$ evaluation protocol . \dag: this model was trained on silhouettes.
    }
    \begin{tabular}{c|cc|cc|cc|cc|cc}
        \hline
        \multirow{2}{*}{}&\multirow{2}{*}{Methods} & \multirow{2}{*}{Venue} & \multicolumn{2}{c|}{NTU RGB-AB} & \multicolumn{2}{c|}{ PKU MMD-AB} &   \multicolumn{2}{c|}{Charades-AB} & \multicolumn{2}{c}{ACC-MM1-Activities}  \\
        & & & Rank 1 & mAP& Rank 1 & mAP& Rank 1 & mAP& Rank 1 & mAP\\ 
        \hline
        \multirow{4}{*}{Image}& CAL \cite{gu2022clothes}&CVPR22 & 70.31 & 24.08& 78.31& 43.43 & 40.13& 21.23 &67.33 &38.21\\
        &PSTR \cite{cao2022pstr}&CVPR22 &68.34& 32.54& 77.98& 41.23& 35.12& 20.32& 53.46& 30.18\\
        &SCNet \cite{guo2023SCNet}&ACM MM23 &68.82& 26.31& 73.91& 39.65& 27.42& 17.61& 55.38& 32.42\\
        &AIM \cite{yang2023good}&CVPR23 &72.79& 30.21 &\underline{79.22}& 44.90& 35.56& \underline{26.36}& 66.81& 38.14\\
        \hline
        \multirow{7}{*}{Video}&TSF \cite{jiang2020rethinking}&AAAI20&67.81& 26.88& 71.61& 33.22& 30.21& 18.29& 41.31& 21.43\\
        &VKD \cite{porrello2020robust}&ECCV20 &66.33& 31.46& 72.19& 34.34& 31.89& 18.81& 51.26& 22.16\\ 
        &BiCnet-TKS \cite{hou2021bicnet}&CVPR21 &69.13& 30.21& 77.13& 33.32& 38.33& 23.34& 58.41& 30.21\\
        &STMN \cite{eom2021video}&ICCV21&70.21& 30.13& 71.53& 42.21& 33.89& 20.81& 57.61& 37.61\\
        &PSTA \cite{wang2021pyramid}& ICCV21&65.13& 31.42& 72.43& \underline{47.42}& 38.72& 24.84& 67.31& 37.33\\
        &SINet \cite{bai2022salient}&CVPR22 &66.21& 27.81& 74.11& 26.21& 37.31& 21.90& 61.32& 36.41\\
        &Video-CAL \cite{gu2022clothes}&CVPR22&\underline{73.31}& \underline{31.73}& 77.34& 45.72& \underline{41.50}& 25.81& \underline{67.48}& \underline{38.23}\\
        \hline
        \multirow{3}{*}{Baselines}& GaitGL \cite{lin2021gait} \dag &-&57.04&27.13&61.22& 27.84& 14.51& 4.85& 35.13& 16.31\\
        &ResNet3D-50 \cite{he2016deep}& - &62.80 & 23.52&65.12&29.41&27.35&14.89&39.89&19.83\\
        & MViTv2 \cite{li2022mvitv2}&-&59.27&21.38&61.40&25.31&21.89&12.79&37.31&17.80\\
         \hline
         &ABNet (ours) &- &\textbf{77.01}&\textbf{37.64}&\textbf{81.44}&\textbf{51.79}&\textbf{44.82}&\textbf{28.78}&\textbf{68.31}&\textbf{38.83}\\
         \hline    
    \end{tabular}
    \label{tab:sota_ca}
\end{table*}

\begin{table}[t]
    \centering
    \caption{\textbf{\textit{Ablation studies}} of each component of ABNet on NTU
    RGB-AB on cross activity evaluation protocol}
    \vspace{-.3cm}
    \begin{tabular}{p{0.4cm}p{0.4cm}p{0.4cm}p{0.4cm}|cc|cc}
         \hline
         \multirow{2}{*}{B/L} & \multirow{2}{*}{K/D} & \multirow{2}{*}{A/P} & \multirow{2}{*}{F/D} & \multicolumn{2}{c|}{View$^+$} & \multicolumn{2}{c}{View$^-$}\\
         &&&&R@1 & mAP& R@1 & mAP\\
         \hline
         \checkmark &&&& 62.80 & 23.52 & 61.71 & 21.41\\
         \checkmark&\checkmark & & &  66.90 & 23.94& 63.03 & 22.01\\
         \checkmark && \checkmark &  & 66.24 &23.81&	64.61 & 22.48 \\
          \checkmark&\checkmark & \checkmark & & 69.21 &31.01& 66.41 & 30.43\\
         \checkmark&\checkmark & & \checkmark & 74.33 &33.79&	72.85 & 31.68\\
          \hline
         \checkmark&\checkmark & \checkmark & \checkmark & \textbf{77.01} &\textbf{37.64}&	\textbf{76.43} & \textbf{36.14}\\
         \hline
    \end{tabular}
    \label{tab:ablation_ca}
\end{table}

\begin{figure*}
    \centering
    \includegraphics[width=.95\linewidth,height=8cm]{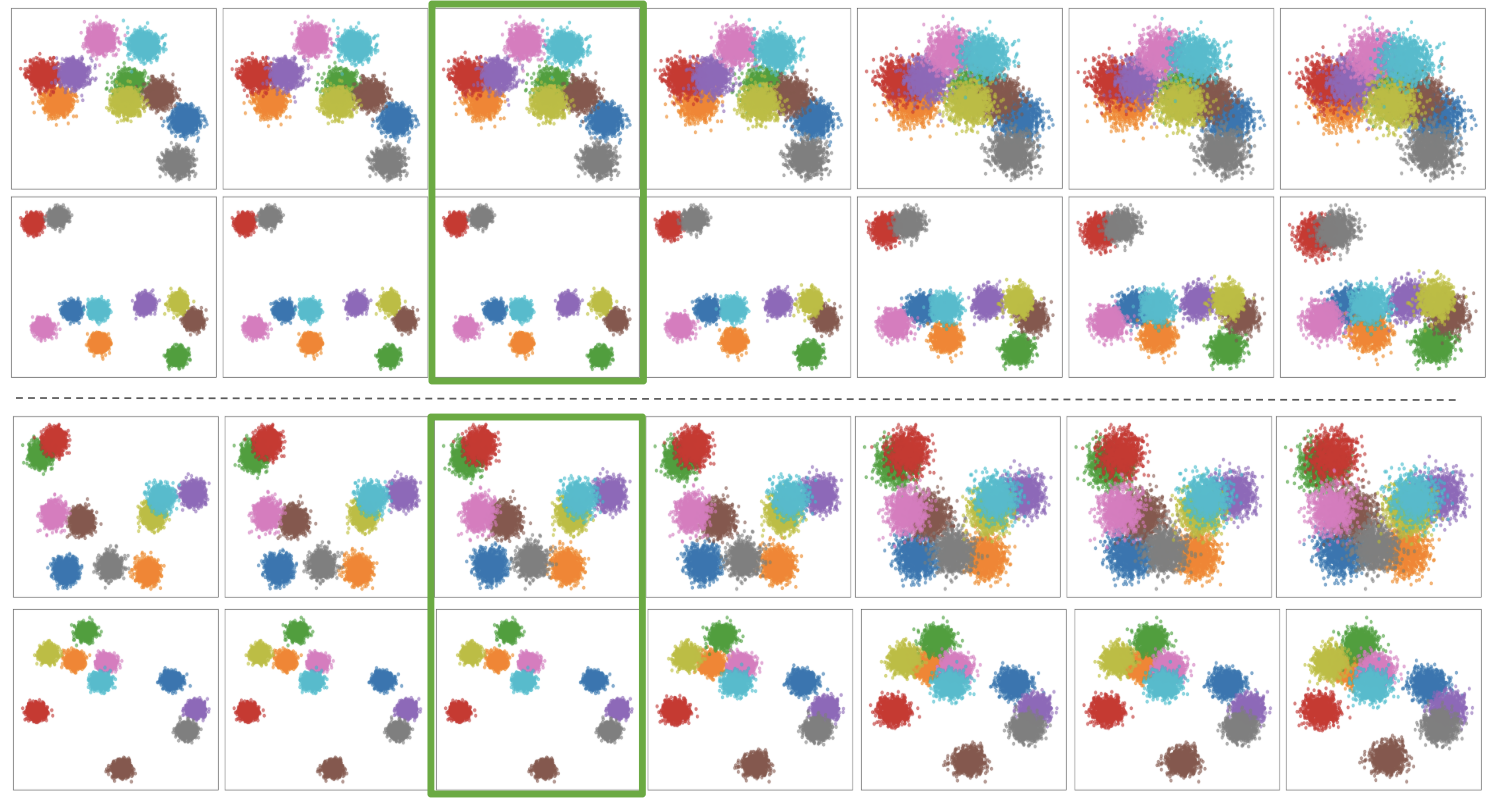}
    \caption{\textbf{\textit{Effect of distortion on feature space.}} The t-SNE plots illustrate the impact of varying distortion amount $\alpha \in 
[200,225,250,275,300,325,350]$ on biometrics (top) and appearance (bottom) features of ABNet for ten random identities for Charades-AB (top) and BRIAR-BGC3 (bottom). As $\alpha$ increases from left to right, the optimal results occur at $\alpha = 250$ (shown in square) where biometrics changes while appearance remains consistent. Beyond $\alpha = 250$, appearance gets distorted too, making it unsuitable for disentanglement}
    \label{fig:tsne-others}
\end{figure*}

\noindent
\textbf{Effect of distortion:} Table \ref{tab:distortion_ca} reports the effect of distortion on cross-activity evaluation protocol on the NTU RGB-AB dataset which is an extension of Table \ref{tab:distortion} of the main paper. Figure \ref{fig:tsne-others} illustrates the t-SNE plots of the biometrics and appearance feature space for ten individuals from two of the challenging datasets; Charades-AB and BRIAR-BGC3. It is observed from this figure that the effect of $\alpha$ is consistent and our choice of $\alpha$ is applicable for even these challenging datasets as well. Figure \ref{fig:distortion_zoom} illustrates the qualitative samples of the effect of distortion. In Figure \ref{fig:distortion_zoom}, we zoomed in the face portion of the actor for better visualization and it is observed that as $\alpha$ increases gradually, the distortion amount increases and by $\alpha=300$, the sample is so distorted that it becomes unsuitable for our purpose.

\begin{table}[t!]
    \centering
    \caption{\textbf{\textit{Effect of distortion}} on model performance for NTU
RGB-AB on the cross activity evaluation protocol}
    \vspace{-.3cm}
    \begin{tabular}{c|cc|cc}
         \hline
         \multirow{2}{*}{Distortion amount}&\multicolumn{2}{c|}{View$^+$} & \multicolumn{2}{c}{View$^-$} \\
         & R@1 & mAP& R@1 & mAP \\
         \hline
        $\alpha=200$&	75.91&37.04&	75.12&35.83\\
        \textbf{$\alpha=250$}& \textbf{77.01} &\textbf{37.64}&	\textbf{76.43} & \textbf{36.14}\\
        $\alpha=300$& 72.70&29.01&	71.03&28.94\\
        \hline 
    \end{tabular}
    \label{tab:distortion_ca}
\end{table}

\begin{table}[t!]
    \centering
    \caption{\textbf{\textit{Effect of face restriction}} on model performance for NTU
RGB-AB on cross activity evaluation protocol}
    \vspace{-.3cm}
    \begin{tabular}{c|cc|cc}
         \hline
         \multirow{2}{*}{Face Restricted}& \multicolumn{2}{c|}{View$^+$} & \multicolumn{2}{c}{View$^-$} \\
         & R@1 & mAP& R@1 & mAP \\
         \hline
         Yes & 77.01 &37.64&	76.43 & 36.14\\
        No& 77.70&39.01&	76.98&38.84\\
        \hline 
    \end{tabular}
    \label{tab:face_blur_ca}
\end{table}

\begin{table}[t!]
    \centering
    \caption{\textbf{\textit{Activity recognition}} performance of different datasets on ABNet.  x-sub and x-view respectively denote cross-subject and cross-view evaluation protocols for its corresponding dataset, if applicable.}
    \begin{tabular}{c|cc}
        \hline
         Dataset & x-sub&x-view\\
         \hline
         NTU RGB-AB & 88.71&89.50\\
         PKU MMD-AB &  91.42&94.21\\
         Charades-AB & 41.31&-\\
         ACC-MM1-Activities & 71.08&-\\
         BRIAR-BGC3&79.31&-\\
         \hline
    \end{tabular}
    \label{tab:activity}
\end{table}

\begin{table*}[t!]
    \centering
    \caption{\textbf{\textit{Choice of Backbone.}} Performance comparison of different backbones on NTU RGB-AB}
    \vspace{-.3cm}
    \begin{tabular}{c|c|cccc|cccc}
         \hline
         \multirow{3}{*}{Network}&\multirow{3}{*}{Backbone}& \multicolumn{4}{c|}{Same activity} & \multicolumn{4}{c}{Cross activity}\\
         && \multicolumn{2}{c}{View$^+$} & \multicolumn{2}{c|}{View$^-$} &\multicolumn{2}{c}{View$^+$} & \multicolumn{2}{c}{View$^-$}\\
         && R@1 & mAP& R@1 & mAP & R@1 & mAP& R@1 & mAP\\
         \hline
        \multirow{3}{*}{Teacher} & \textbf{GaitGL}\cite{lin2021gait}&	\textbf{61.51}&\textbf{28.89}&	\underline{57.78}&\textbf{26.78}&	\underline{57.04}&\textbf{27.13}&	\underline{55.80}&\textbf{26.41}\\
        &GaitPart\cite{fan2020gaitpart}&	54.79& 16.73&	53.93&15.91&	52.18&15.01&	46.89&13.84\\
        &GaitBase\cite{fan2023opengait}&	60.21&28.02&	\textbf{59.04}&26.76&	\textbf{59.90}&26.31&	\textbf{57.91}&25.96\\
        \hline
        \multirow{5}{*}{Student} & MViT v2\cite{li2022mvitv2}&	63.87&26.41&	61.01&\textbf{23.81}&	59.27&21.38&	59.16&20.01\\
        & ViViT\cite{arnab2021vivit} &58.81&20.41&	57.10&16.42&	57.30&12.41&	52.01&9.68\\
        & Swin\cite{liu2021swin} & 59.20&21.68&	58.41&19.41&	58.70&16.91&	54.31&11.47\\
        \cline{2-10}
        & \textbf{ResNet3D-50}\cite{he2016deep} & \textbf{64.23}&\textbf{26.89}&	\textbf{62.10}&\underline{22.45}&	\textbf{62.80}&\textbf{23.52}&	\textbf{61.71}&\textbf{21.41}\\
        & ResNet3D-34\cite{he2016deep} & 63.90&25.93&	60.45&21.87&	60.21&22.74&	59.79&20.47\\
        \hline 
    \end{tabular}
    \label{tab:tbackbone}
\end{table*}

\begin{figure}[t!]
    \centering
    \includegraphics[width=.95\linewidth]{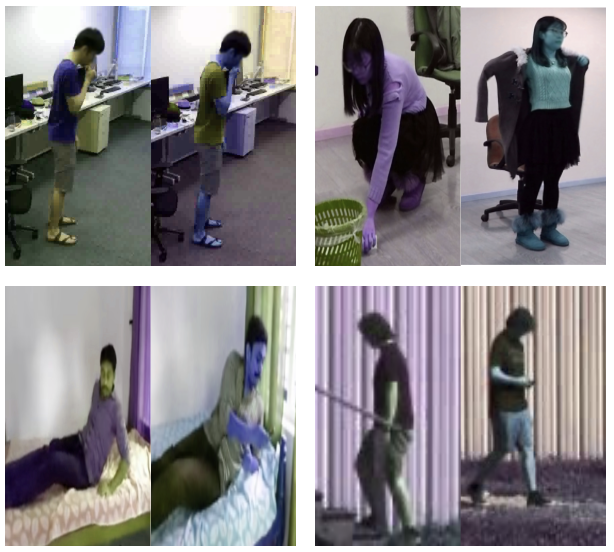}
    \caption{\textbf{\textit{Dataset samples.}} Here each two samples show different values of hue shifting for the same video of NTU RGB-AB (top-left), PKU MMD-AB (top-right), Charades-AB (bottom-left) and ACC-MM1-AB (bottom-right). All the samples have their faces blurred.}
    \label{fig:data_samp}
    \vspace{-.1cm}
\end{figure}

\begin{figure*}[t!]
    \centering
    \includegraphics[width=0.49\linewidth,height=4.1cm]{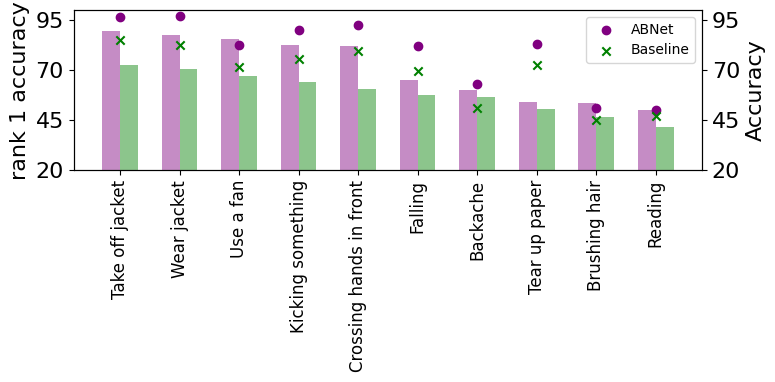}
    \includegraphics[width=0.49\linewidth]{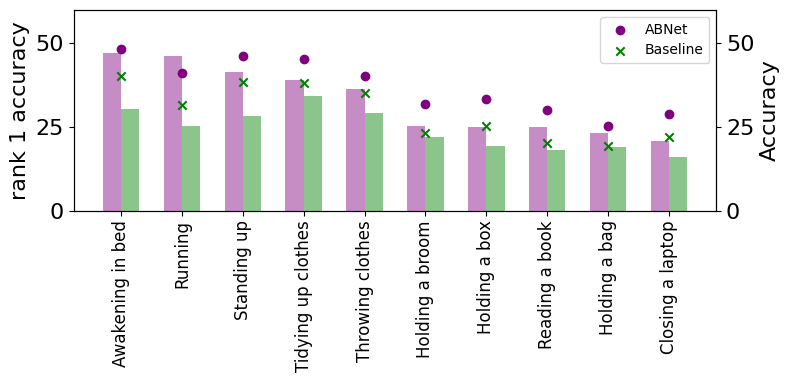}
    \includegraphics[width=0.49\linewidth]{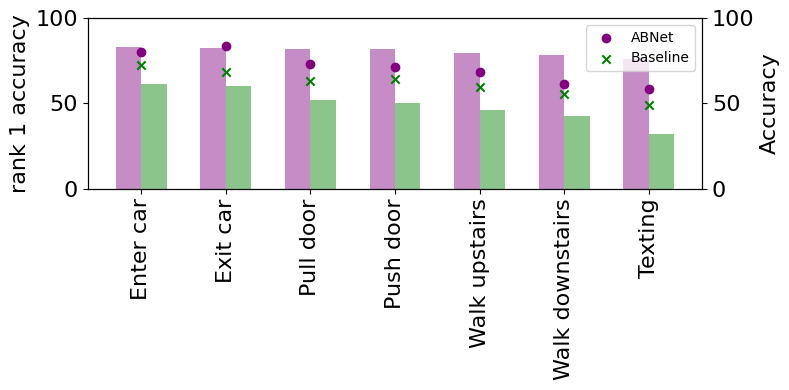}
    \includegraphics[width=0.49\linewidth]{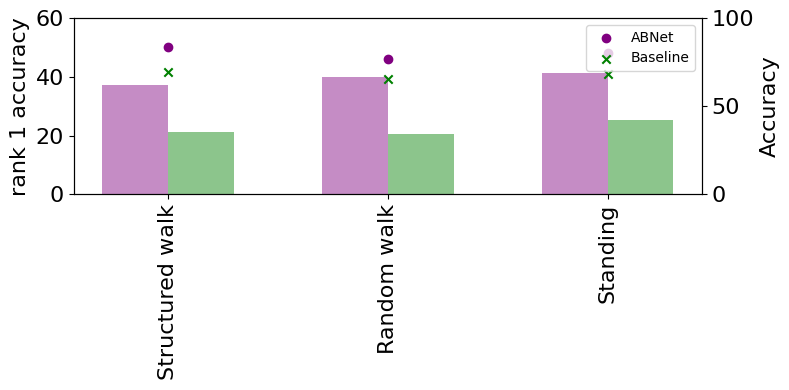}
    \caption{\textbf{\textit{Performance analysis across activities.}} The bar plot on
left axis shows rank 1 identification accuracy for given activity of
ABNet against baseline PKU MMD-AB (top-left), Charades AB (top-right), ACC-MM1-Activities (bottom-left) and BRIAR-BGC3 (bottom-right) datasets. The scatter plot with markers on right axis shows activity recognition accuracy for corresponding classes.}
    \label{fig:act_pku_ch}
\end{figure*}

\begin{figure*}[t!]
    \centering
    \includegraphics[width=.49\linewidth]{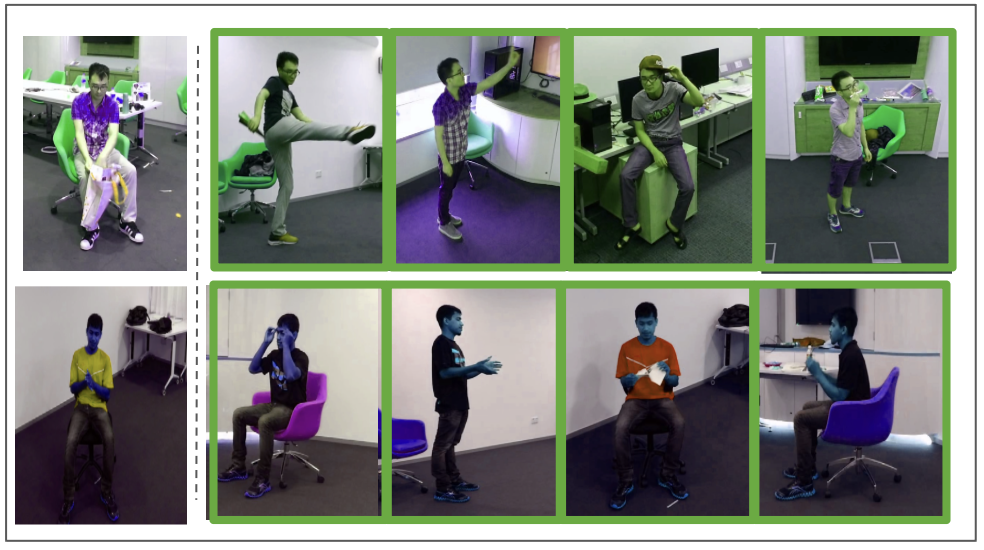}
    \includegraphics[width=.49\linewidth]{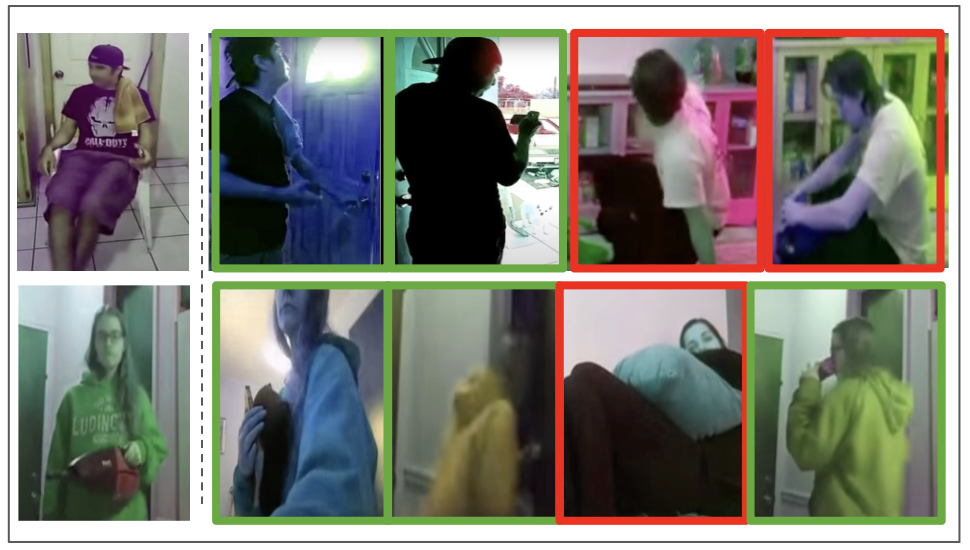}
    \includegraphics[width=.49\linewidth]{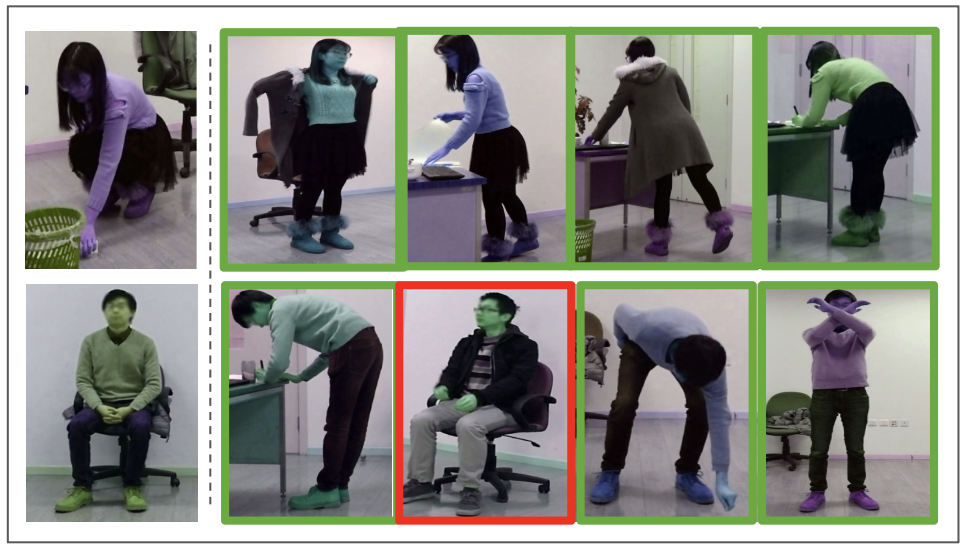}
    \includegraphics[width=.49\linewidth]{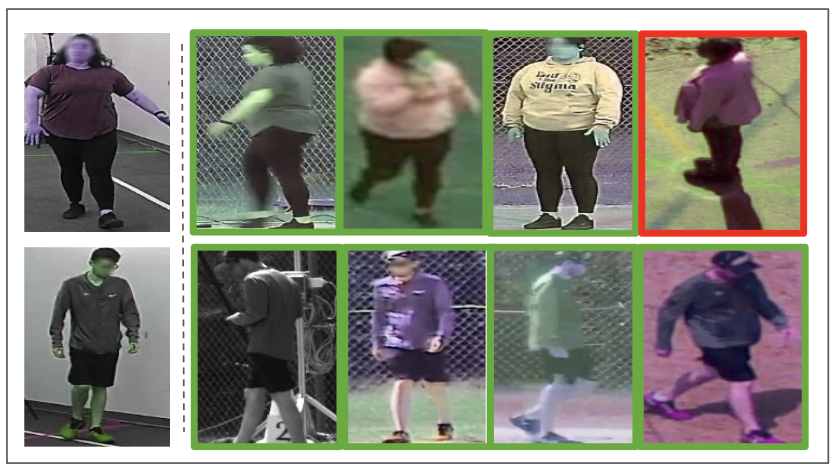}
    \caption{\textbf{\textit{Top 4 rank retrieval}} samples for ABNet on NTU RGB-AB (top-left), Charades-AB (top-right), PKU MMD-AB (bottom-left) and BRIAR-BGC3 (bottom-right). The left most columns for each dataset samples hold the probe samples and the following four columns to that probe are its retrieval list.  Accurate retrieval is shown with green box and inaccurate with red.}
    \label{fig:qual_supp}
\end{figure*}

\noindent
\textbf{Effect of face restriction on cross-activity evaluation protocol} is reported in Table \ref{tab:face_blur_ca} on the NTU RGB-AB dataset. Similar to the results reported in the main paper, even in case of the cross-activity evaluation protocol, the model performance remains stable even when faces are restricted showing the learning of non-facial cues across cross-activity evaluation protocol.

\noindent
\textbf{Choice of backbone.} The performance comparison of different backbone networks is shown in Table \ref{tab:tbackbone}, where the backbone model takes the silhouette/RGB video frames as input respectively for the teacher/student network for the task of person identification. Here this experiment is run only on the baseline where none of the modification components are present. This selection of backbones ensures that the teacher network contributes its expertise to the specific task it is designed for in the student network. Moreover, similar to existing recent work \cite{gu2022clothes,hou2021bicnet,cao2022pstr,yang2023good} in person identification, in our case also CNN based backbones outperform transformer based ones. From this experiment, we pick the best performing backbone for both networks.

\noindent
\textbf{Performance of action recognition:} Table \ref{tab:activity} reports the performance of ABNet on activity recognition results for different datasets. Here the reported evaluation metric is accuracy on cross-subject and cross-view evaluation protocol. NTU RGB-AB and PKU MMD-AB are evaluated on these two protocols, however, since there is no explicit view information for rest of the three datasets, the accuracies are reported in terms of cross-subject because the test and train split contains mutually exclusive actors/subjects.\\
Figure \ref{fig:act_pku_ch} compare ABNet and the baseline across the top five best and bottom five worst performing activities in person identification for PKU MMD-AB and Charades (top row). The bottom row shows person identification performance across all 7 activities of the ACC-MM1-Activities dataset and all 3 activities of BRIAR-BGC3 dataset. It is observed that activities with minimal overall body movement pose greater challenges for individual identification, whereas more overall body movement contribute to higher person identification accuracy. This highlights the significance of incorporating activity prior in our model. Moreover, it also emphasizes the importance of activity cues demonstrating the efficacy of our joint training approach in effectively learning such cues.

\noindent
\textbf{Accuracy of silhouette extractor and effectiveness of silhouettes:} The accuracy of the silhouette extraction process will indeed affect model's performance and to explore that we perform an experiment using Grounded-SAM \cite{ren2024grounded} which is an open-world segmentation model. The results are reported on a small subset (10 action classes) of the NTU-RGB-AB dataset on the same activity View$^+$ setting in Table \ref{tab:sils}.
It is observed that with Grounded-SAM as silhouette extractor the performance does go up, which can be attributed to it being an open-world model and thus being more robust. Similarly, a $\mathbf{3\%}$ rank 1 accuracy gain is seen in case of a small subset of the Charades-AB dataset when using Grounded-SAM as opposed to Mask2Former. Nevertheless, even with a weaker silhouette extractor our model still performs well and since this extraction process is not part of the inference stage, training the model with a better silhouette extractor will provide some benefits.
\begin{table}
    \centering  
    \caption{\textbf{\textit{Performance with varying silhouette extractors}}}
    \begin{tabular}{c|cc}
        \hline
         Silhouette extractor & Rank 1 & mAP \\
         \hline 
         Mask2Former \cite{cheng2021mask2former} & 85.2&87.3\\
         Grounded-SAM \cite{ren2024grounded}&87.8&88.5\\
         \hline
    \end{tabular}
    \label{tab:sils}
    \vspace{-.5cm}
\end{table}
The main motivation behind using silhouette features is to distill appearance-less knowledge, e.g. purely biometrics information that not only contains gait; but also pose, body shape, structure etc information to aid disentanglement. The recognition performance of the two decoupled features is reported in Table \ref{tab:recog} for the Charades-AB dataset. The huge performance gap between the biometrics and non-biometrics features shows that the non-biometrics features do not have meaningful information to perform person identification; essentially proving the effectiveness of the disentanglement process. To demonstrate the effectiveness of using silhouette features in our method, we distort the silhouettes and distill that knowledge to the biometrics features, which resulted in a huge performance drop (about $\mathbf{24\%}$) (row 3 of Table \ref{tab:recog}). This shows that even in case of activities beyond walking, the silhouette-based biometrics features contribute to a great extent in accurate recognition. We specifically select the Charades-AB dataset for this experiment as it is a real-world dataset encompassing a diverse range of appearance variations. 

\begin{table}
    \centering 
    \caption{\textbf{\textit{Performance of disentangled features}}}
    \begin{tabular}{c|cc}
        \hline
         Feature&Rank 1 &mAP  \\
         \hline
         Biometrics&\textbf{45.8}&\textbf{31.6}\\
         Non-biometrics&2.8&0.4\\
         Biometrics w/ distorted sils &21.4 &10.5\\
         \hline
    \end{tabular}
    \label{tab:recog}
\end{table}

\section{Qualitative analysis}
\label{sec:qualitative_supp}
Figure \ref{fig:data_samp} illustrates examples of different values of hue-shifting, from which it can be observed that the color profile for each frame is distinct from the other. 
Figure \ref{fig:qual_supp} illustrates the top 4 rank retrieval results for a given probe for NTU RGB-AB, PKU MMD-AB and Charades-AB datasets. Some of the failure cases is seen for having difficulty performing accurate retrieval due to the absence of a lot of overall body movement (e.g. probe activity is sitting in first sample of Charades-AB and second sample of PKU MMD-AB). Moreover, another failure case is seen in case of the second sample of Charades-AB which shows the inherent challenges present in the dataset, e.g. data quality, no standard way of performing an activity etc. Despite these challenges, from the figure it is observed that accurate retrieval is done in most cases irrespective of view-point, activity and appearance, which shows the effectiveness of ABNet.

\end{document}
\typeout{get arXiv to do 4 passes: Label(s) may have changed. Rerun}

%% file: preamble.tex
\usepackage[dvipsnames]{xcolor}
\usepackage{multirow}
\usepackage{tabularx}
\usepackage{tablefootnote}
\usepackage{float}
\usepackage{amsmath}

\usepackage{xr}
\usepackage[accsupp]{axessibility} 